\documentclass[11pt]{article}
\usepackage[final]{acl}
\usepackage{times}
\usepackage{latexsym}
\usepackage{amsmath}
\usepackage{booktabs}
\usepackage{multirow}
\usepackage{comment}
\usepackage{graphicx}
\usepackage[T1]{fontenc}
\usepackage[utf8]{inputenc}
\usepackage{microtype}
\usepackage{inconsolata}
\usepackage{graphicx}
\usepackage{subcaption}
\usepackage[table]{xcolor}
\usepackage{amssymb}
\definecolor{sigpos}{RGB}{226,240,217}
\definecolor{signeg}{RGB}{252,228,214}

\definecolor{gaplow}{RGB}{255,248,232}
\definecolor{gapmid}{RGB}{255,232,191}
\definecolor{gaphigh}{RGB}{255,209,168}

\definecolor{belowlight}{RGB}{221,235,247}
\definecolor{belowstrong}{RGB}{189,215,238}
\definecolor{abovelight}{RGB}{253,228,205}
\definecolor{abovestrong}{RGB}{248,203,173}

\definecolor{aucchance}{RGB}{242,242,242}
\definecolor{aucweak}{RGB}{234,244,227}
\definecolor{aucmedium}{RGB}{213,235,199}
\definecolor{aucstrong}{RGB}{183,223,167}

\hypersetup{
    colorlinks=true,
    linkcolor=blue
}

\title{Who Pays for Efficient Inference? Confidence-Based Small-to-Large LLM Routing Across Tasks and Languages}

\author{
\textbf{Md Sabbir Hossen\textsuperscript{1}} \hspace{5mm}
\textbf{Anichur Rahman\textsuperscript{2}} \hspace{5mm}
\textbf{Pabon Shaha\textsuperscript{1}} \hspace{5mm}
\textbf{Airin Afroj Aishi\textsuperscript{2}} \\
\textbf{Andrew H. Sung \textsuperscript{3}} \hspace{5mm}
\textbf{Md Shohel Rana\textsuperscript{2}} \vspace{2mm}
\\
\textsuperscript{1}Bangladesh University, Bangladesh \hspace{5mm}
\textsuperscript{2}Georgia Southern University, USA \\
\textsuperscript{3}University of Southern Mississippi, USA \\
\small{
\textbf{Correspondence:}
\href{mailto:ar36248@georgiasouthern.edu}{ar36248@georgiasouthern.edu};
\href{mailto:mrana@georgiasouthern.edu}{mrana@georgiasouthern.edu}
}}
\begin{document}
\maketitle
\maketitle

\begin{abstract}
\vspace{-0.8mm}
Small-to-large language model routing can reduce inference cost by escalating difficult inputs while retaining easier ones on a smaller model. Its success, however, depends on identifying examples that benefit from escalation, and the reliability of confidence across multilingual tasks remains unclear. We present a systematic evaluation of confidence-based routing across three model families, four languages, and three task types, covering 72 model, dataset, language, and scale configurations and 42,288 predictions. We compare global confidence routing with random routing, equal-quota confidence routing, and an oracle upper bound. Confidence identifies beneficial escalations on the Belebele reading comprehension and SIB-200 topic classification datasets, with mean correction-detection AUCs of 0.612 and 0.705. At 50\% escalation, it improves over random routing by 1.40--4.67 percentage points on Belebele and 2.54--5.63 points on SIB-200. In contrast, the mean AUC on the MMLU-ProX-Lite knowledge and reasoning dataset is only 0.510, and routing provides no consistent improvement over random selection because small models often remain highly confident while incorrect. Global confidence routing also creates language escalation gaps up to 30.33 percentage points, whereas equal-quota routing removes this disparity with little aggregate accuracy cost. These results show that confidence routing is task-dependent and can allocate computation unevenly across languages. Code is available at \href{https://github.com/sabbir5622r/multilingual-llm-routing}{\texttt{GitHub}}.

\end{abstract}

\section{Introduction}
\label{sec:introduction}

Large language models provide strong performance across many natural language processing tasks \citep{NEURIPS2020_1457c0d6,JMLR:v24:22-1144} but require substantial computational resources \citep{pmlr-v202-sheng23a,10.1145/3600006.3613165}. Smaller models are less expensive to deploy \citep{abdin2024phi-,pmlr-v235-liu24ce}, although their more limited capacity can result in lower performance \citep{kaplan2020scalinglawsneurallanguage,NEURIPS2022_c1e2faff}. Small-to-large routing addresses this trade-off by processing each query with a smaller model and escalating selected queries to a larger model \citep{ding2024hybridllm,chen2024frugalgpt}. An effective router can therefore preserve much of the larger model's performance while reducing how frequently it is used \citep{agarwalS2024_ecda225c}.

\begin{figure}[t]
    \centering
    \includegraphics[width=0.48\textwidth]{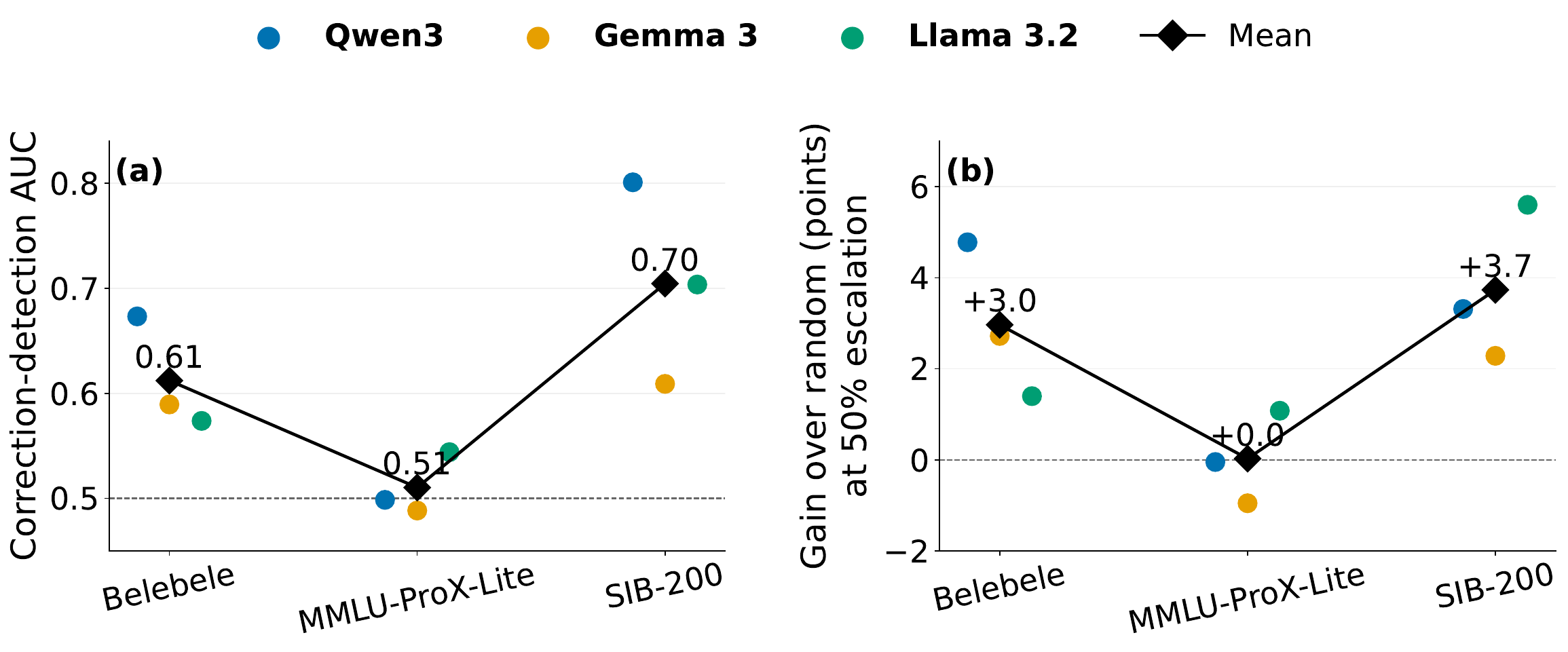}
    \caption{\textbf{Confidence routing is task-dependent.} Panel (a) reports correction-detection AUC, and (b) reports accuracy gain over random routing at 50\% escalation. Colored points represent model families, while black diamonds denote their mean. Dashed lines indicate chance-level AUC and zero gain over random routing.}
    \vspace{-2mm}
    \label{fig:introduction_teaser}
\end{figure}

A central challenge is determining which queries should be escalated \citep{ICLR2024_11f5520d,ICLR2024_5de11e93}. Existing approaches estimate query difficulty, learn model preferences, or use confidence signals produced by the smaller model \citep{ding2024hybridllm,ongICLR2025_5503a7c6,shen-etal-2025-sater}. Confidence is particularly practical because it can be derived directly from the smaller model's outputs without requiring a separate routing model \citep{geng-etal-2024-survey,zhang2025leveraginguncertaintyestimationefficient}. However, this strategy assumes that low confidence identifies small-model errors that the larger model can correct \citep{pmlr-v306-miao26b,luo2026routelmtlearnedsamplerouting}. Language models can remain highly confident while incorrect \cite{geng-etal-2024-survey}, and confidence may not reflect the benefit of consulting a larger model \citep{luo2026routelmtlearnedsamplerouting}. This concern is especially relevant in multilingual settings, where calibration varies across languages \citep{ahuja-etal-2022-calibration} and global confidence ranking may therefore allocate escalation unevenly. \textbf{This motivates our primary research question:} \textit{How reliably does small-model confidence identify multilingual inputs that benefit from escalation, and how do task type, model family, and language affect the resulting accuracy, computation, and allocation trade-offs?} Figure~\ref{fig:introduction_teaser} previews how the reliability and benefit of confidence routing vary across tasks.

To answer this question, we conduct a controlled evaluation of Qwen3, Gemma~3, and Llama~3.2 using smaller and larger models across English, Bangla, Hindi, and Urdu. We evaluate reading comprehension, knowledge and reasoning, and topic classification using Belebele, MMLU-ProX-Lite, and SIB-200, resulting in 72 model, dataset, language, and scale configurations. We compare global confidence routing with random routing, equal-quota confidence routing, and an oracle upper bound through accuracy and escalation curves, correction detection, confidence quintiles, and language-level allocation analysis. This design provides a systematic multilingual assessment of when confidence predicts beneficial escalation, distinguishes routing utility from ordinary prediction uncertainty, and evaluates equal-quota routing as an allocation-aware alternative.

 \section{Related Work}
\label{sec:related_work}

\subsection{LLM Routing and Cascading}

LLM routing reduces inference cost by assigning queries to models with
different capability and cost levels. \citet{ding2024hybridllm} introduced Hybrid LLM, which predicts
query difficulty to route inputs between smaller and larger models.
Related frameworks used cascaded API calls or small-model
self-verification to balance response quality and computational
cost~\citep{chen2024frugalgpt, agarwalS2024_ecda225c}. \citet{ongICLR2025_5503a7c6} subsequently trained routers from
preference data, while~\citet{pmlr-v267-dekoninck25a}
formally unified routing and cascading. More recent methods jointly
select models and sampling budgets or learn explicit confidence signals
for escalation~\citep{ding2025bestrouteadaptivellmrouting,pmlr-v267-chuang25b, chuang2025confidentseekstrongerexploring}.

RouterBench introduced a standardized framework for comparing multi-LLM
routers~\cite{hu2024routerbenchbenchmarkmultillmrouting}. Its larger successor, LLMRouterBench,
showed that several sophisticated routers do not consistently
outperform simple baselines~\cite{li2026llmrouterbenchmassivebenchmarkunified}. Other analyses
similarly found that routing effectiveness can vary substantially
across tasks, query categories, and evaluation
settings~\citep{srivatsa-etal-2024-harnessing,kassem-etal-2026-robust}.

\subsection{Confidence and Multilingual Calibration}

Confidence estimation provides a natural signal for deciding whether a
prediction should be accepted or escalated. \citet{geng-etal-2024-survey} organized existing methods into
probability-based, consistency-based, self-evaluation, and verbalized
confidence approaches. \citet{jiang-etal-2021-know} found
that generative language models were poorly calibrated on
question-answering tasks. Subsequent studies showed that LLMs possess some ability to evaluate their answers and recognize the boundaries of their knowledge, although this ability does not transfer reliably across tasks~\citep{kadavath2022languagemodelsmostlyknow,yin2023largelanguagemodelsknow}.

Confidence reliability is especially challenging in multilingual settings. \citet{ahuja-etal-2022-calibration} observed
greater miscalibration in low-resource languages and languages
typologically distant from English. ~\citet{yang2024calibrationmultilingualquestionanswering} likewise found that multilingual QA
models were generally less well calibrated outside English, although
multilingual calibration examples could partially reduce these errors.
These findings question whether confidence can serve as a universally
reliable routing signal.

\subsection{Multilingual Evaluation and Disparity}

\citet{ahuja-etal-2024-megaverse} demonstrated substantial
performance variation across languages, models, and task types in a
large-scale multilingual evaluation. \citet{chang-etal-2024-multilinguality} further showed that multilingual
training can benefit low-resource languages under limited data but
become harmful when model capacity is constrained. Related fairness
studies found that cross-language disparities depend jointly on the
task, model, language, and available
supervision~\citep{hu-etal-2023-systematic,cabello-piqueras-sogaard-2022-pretrained}. \citet{hu-etal-2025-quantifying} additionally showed that higher
aggregate performance does not necessarily imply smaller language
disparities. \citet{singh-etal-2025-global} demonstrated that
translated benchmarks can preserve English-centric cultural assumptions
and alter multilingual model rankings.

\section{Experimental Design}
\label{sec:experimental_setup}

\subsection{Datasets and Languages}
\label{sec:datasets}

We evaluate three multilingual benchmarks: Belebele for reading comprehension \citep{bandarkar-etal-2024-belebele}, MMLU-ProX-Lite for knowledge and reasoning \citep{xuan-etal-2025-mmlu}, and SIB-200 for topic classification \citep{adelani-etal-2024-sib}. Table~\ref{tab:datasets} summarizes their evaluated splits and sample counts per language. Our evaluation includes English, Bangla, Hindi, and Urdu, covering four scripts: Latin, Bengali, Devanagari, and Arabic. Across six models, the evaluation contains 42,288 predictions and 21,144 paired small-to-large comparisons. Additional dataset and preprocessing details are provided in Appendix~\ref{app:data}. \vspace{-1mm}

\begin{table}[htbp]
    \centering
    \scriptsize
    \setlength{\tabcolsep}{3.8pt}
    \begin{tabular}{lllr}
        \toprule
        Dataset & Task & Split & $N$/language \\
        \midrule
        Belebele & Reading comprehension & Test & 900 \\
        MMLU-ProX-Lite & Knowledge and reasoning & Val.\ + test & 658 \\
        SIB-200 & Topic classification & Test & 204 \\
        \bottomrule
    \end{tabular}
    \caption{Datasets used in this study and the number of examples evaluated per language.}
    \label{tab:datasets}
\end{table}

\subsection{Model Families}
\label{sec:models}

We evaluate three instruction-tuned model families: Qwen3~\cite{yang2025qwen3technicalreport}, Gemma~3~\cite{gemmateam2025gemma3technicalreport}, and Llama~3.2~\cite{grattafiori2024llama3herdmodels}. For each family, we select a smaller model containing 1B to 1.7B parameters and a larger model containing 3B to 4B parameters. Pairing models from the same family provides a controlled comparison of model capacity while reducing variation arising from differences in architecture, tokenizer, and training procedure. Including three distinct families allows us to examine whether routing behavior generalizes beyond a particular model family.

Routing is performed only between models belonging to the same family. A query initially processed by the smaller Qwen3 model, for example, may only be escalated to the larger Qwen3 model. This design isolates the effect of model scale within each family and avoids introducing additional variation through cross-family routing. Table~\ref{tab:models} summarizes the model pairs used in our experiments.

\begin{table}[htbp]
    \centering
    \scriptsize
    \setlength{\tabcolsep}{7pt}
    \begin{tabular}{lll}
        \toprule
        Family & Small model & Large model \\
        \midrule
        Qwen3 &
        \texttt{Qwen3-1.7B} &
        \texttt{Qwen3-4B} \\
        Gemma 3 &
        \texttt{gemma-3-1b-it} &
        \texttt{gemma-3-4b-it} \\
        Llama 3.2 &
        \texttt{Llama-3.2-1B-Instruct} &
        \texttt{Llama-3.2-3B-Instruct} \\
        \bottomrule
    \end{tabular}
    \caption{Small and large models used for each family and paired within the same family for routing}
    \label{tab:models}
\end{table}

\subsection{Multiple-Choice Scoring}
\label{sec:choice_scoring}

We evaluate all tasks using deterministic next-token multiple-choice scoring. Each example is converted into a prompt containing its input and candidate answers. We apply the model family's instruction or chat template; for Qwen3, explicit reasoning mode is disabled. The model is not allowed to freely generate an answer. Instead, we extract the next-token logit associated with each valid symbolic answer label.

Let $\mathcal{Y}_i$ denote the candidate-label set for example $i$, and let $z_{i,c}$ be the next-token logit assigned to candidate $c$. We normalize only over the valid candidates:
\begin{equation}
    p_i(c)
    =
    \frac{\exp(z_{i,c})}
    {\sum_{c' \in \mathcal{Y}_i}\exp(z_{i,c'})}.
    \label{eq:choice_probability}
\end{equation}
The predicted label is
\begin{equation}
    \hat{y}_i
    =
    \arg\max_{c \in \mathcal{Y}_i} p_i(c).
\end{equation}

This constrained scoring procedure avoids variation from decoding and eliminates failures caused by extracting labels from unconstrained generated text.

Our primary confidence signal is the probability margin between the highest- and second-highest-scoring candidates:
\begin{equation}
    m_i
    =
    p_i(c_i^{(1)}) - p_i(c_i^{(2)}),
    \label{eq:confidence_margin}
\end{equation}
where $c_i^{(1)}$ and $c_i^{(2)}$ are the first- and second-ranked candidates, respectively. A smaller margin indicates greater apparent uncertainty and therefore a higher priority for escalation.

\subsection{Implementation and Numerical Precision}
\label{sec:implementation}

All models are evaluated with batch size one, a maximum input length of 4,096 tokens, and random seed 42. We resolve and record the exact Hugging Face revision SHA for every checkpoint to support reproducibility. Predictions are cached at the example level so interrupted evaluations can resume without recomputing completed examples. 

Experiments were conducted using NVIDIA Tesla T4 GPUs. Most checkpoints were evaluated in FP16. However, Gemma 3 4B produced non-finite logits in FP16 on MMLU-ProX-Lite and SIB-200. We therefore re-evaluated these configurations in FP32 across two T4 GPUs. We additionally restricted the output computation to the final token position, which substantially reduced memory use without changing the resulting prediction. A separate numerical check confirmed finite logits and agreement between full-sequence and final-token-only scoring. The Belebele Gemma 3 4B outputs were retained because their original FP16 logits were finite and passed the validation checks.

Because these evaluations involve different numerical precisions and device placements, we use escalation rate, rather than measured wall-clock latency, as the principal cost axis. We consequently interpret the results as accuracy--escalation trade-offs and do not make cross-model latency claims.

We additionally measure system efficiency on 1,200 matched inputs per family using two NVIDIA Tesla T4 GPUs, batch size one, five warm-up examples, and three repetitions. We report latency, throughput, GPU energy, and GPU-hours. Model loading is excluded, while routing overhead is included. Gemma~3 4B uses FP32, whereas the remaining models use FP16. Appendix~\ref{app:system_efficiency} provides the complete protocol.

\section{Routing Methodology}
\label{sec:routing_methodology}

We simulate a cascaded inference system in which every input is first scored by a small model. A routing policy then decides whether to accept the small model's prediction or escalate the input to the corresponding larger model. All routing experiments are computed from paired small- and large-model predictions on the same examples.

\subsection{Routing Objective}
\label{sec:routing_objective}

For example $i$, let
\begin{equation}
    s_i
    =
    \mathbb{I}
    \left[
        \hat{y}^{\,\mathrm{small}}_i = y_i
    \right]
\end{equation}
denote small-model correctness, and let
\begin{equation}
    l_i
    =
    \mathbb{I}
    \left[
        \hat{y}^{\,\mathrm{large}}_i = y_i
    \right]
\end{equation}
denote large-model correctness. A routing decision $r_i \in \{0,1\}$ indicates whether the example is escalated. The resulting correctness is
\begin{equation}
    a_i(r_i)
    =
    (1-r_i)s_i + r_i l_i.
    \label{eq:routed_correctness}
\end{equation}
Thus, the routed system uses the small-model prediction when $r_i=0$ and the large-model prediction when $r_i=1$.

For a set of $N$ examples, routed accuracy is
\begin{equation}
    A(\mathbf{r})
    =
    \frac{1}{N}\sum_{i=1}^{N}a_i(r_i),
\end{equation}
and the realized escalation rate is
\begin{equation}
    \rho(\mathbf{r})
    =
    \frac{1}{N}\sum_{i=1}^{N}r_i.
    \label{eq:escalation_rate}
\end{equation}

At a target budget $b$, a fixed-budget policy selects approximately
\begin{equation}
    K_b = \operatorname{round}(bN)
\end{equation}
examples for escalation. We report complete routing curves for $b \in \{0,0.1,\ldots,1.0\}$ and focus our statistical comparisons on the 25\% and 50\% escalation budgets.

\subsection{Routing Policies}
\label{sec:routing_policies}

We compare four principal routing policies.

\paragraph{Global confidence routing.}
We pool the four languages within each dataset--family pair and rank all examples by the small model's confidence margin in ascending order. At budget $b$, the $K_b$ lowest-margin examples are escalated:
\begin{equation}
    r_i^{\mathrm{conf}}(b)
    =
    \mathbb{I}
    \left[
        i \in
        \operatorname{BottomK}
        \left(
            \{m_j\}_{j=1}^{N},
            K_b
        \right)
    \right].
\end{equation}
This policy represents a practical global router that requires neither ground-truth labels nor a learned routing model. However, it may allocate different escalation rates to different languages if their confidence distributions differ.

\paragraph{Equal-quota confidence routing.}
To enforce allocation parity, we assign approximately the same escalation fraction to every language. For language $\ell$ with $N_\ell$ examples, we select
\begin{equation}
    K_{b,\ell}
    =
    \operatorname{round}(bN_\ell)
\end{equation}
examples with the lowest within-language confidence margins. This produces approximately equal language-level escalation rates by construction. Importantly, equal allocation does not guarantee equal accuracy or equal improvements across languages.

\paragraph{Random routing.}
Random routing uniformly selects $K_b$ examples without using model confidence. We repeat random selection 100 times at each budget and report the mean routed accuracy. This baseline measures the benefit of confidence ranking beyond allocating the same escalation budget arbitrarily.

\paragraph{Oracle routing.}
We define the per-example gain from escalation as
\begin{equation}
    g_i = l_i - s_i.
    \label{eq:oracle_gain}
\end{equation}
Oracle routing ranks examples by $g_i$ and escalates those with the greatest gain. This policy has access to both models' correctness and is therefore not deployable. It provides an upper bound on the value available from perfectly identifying beneficial escalations.

Because a large model can occasionally change a correct small-model answer into an incorrect one, $g_i$ may equal $-1$. Under a fixed-budget formulation, the oracle can therefore become non-monotonic at very high budgets after all beneficial and neutral examples have already been selected. We interpret the oracle as diagnostic headroom rather than as a realistic routing method.

\subsection{Evaluation Metrics and Confidence Diagnostics}
\label{sec:confidence_diagnostics}

We use accuracy as the primary measure of model and routing performance. We report small-model accuracy, large-model accuracy, and routed accuracy, calculated separately for each language and macro averaged across English, Bangla, Hindi, and Urdu. We use the realized escalation rate as the principal cost measure and report accuracy across budgets from 0\% to 100\%, with statistical comparisons focused on the 25\% and 50\% budgets.

A useful confidence score should identify examples on which escalation corrects a small-model error. We define the correction indicator as
\begin{equation}
    c_i
    =
    \mathbb{I}[s_i=0 \land l_i=1].
    \label{eq:correction_indicator}
\end{equation}
We measure how well low confidence predicts $c_i=1$ using the area under the receiver operating characteristic curve, with $-m_i$ as the detection score. An AUC of 0.5 indicates chance-level ranking, whereas a larger value indicates that low-margin examples are more likely to benefit from escalation.

We divide examples into confidence quintiles and measure small-model accuracy and error rate within each quintile. This analysis reveals whether errors are concentrated among uncertain examples or whether the model remains confidently wrong. 

\subsection{Cross-Lingual Allocation Metrics}
\label{sec:fairness_metrics}

For each language $\ell$, let $\rho_\ell$ denote its escalation rate and $A_\ell$ its routed accuracy. We measure allocation disparity using the escalation gap
\begin{equation}
    \Delta_{\mathrm{esc}}
    =
    \max_{\ell}\rho_\ell
    -
    \min_{\ell}\rho_\ell.
    \label{eq:escalation_gap}
\end{equation}
We further report worst-language accuracy
\begin{equation}
    A_{\mathrm{worst}}
    =
    \min_{\ell}A_\ell
\end{equation}
and the best--worst language accuracy gap
\begin{equation}
    \Delta_A
    =
    \max_{\ell}A_\ell
    -
    \min_{\ell}A_\ell.
    \label{eq:accuracy_gap}
\end{equation}

These metrics distinguish allocation parity from outcome parity. In particular, equal-quota routing guarantees a near-zero escalation gap, but it does not necessarily equalize language-level accuracy because the models' initial accuracies and the benefits of escalation can differ by language.

\subsection{Statistical Inference}
\label{sec:statistical_inference}

We estimate uncertainty using 5,000 paired clustered-bootstrap repetitions. Within each dataset--family pair, resampling is performed over example IDs, keeping the corresponding language versions of an example in the same bootstrap cluster. This preserves both the pairing between routing policies and the dependence among parallel multilingual examples.

At the 25\% and 50\% budgets, we estimate paired differences between global confidence routing and (i) random routing and (ii) equal-quota confidence routing. For the random baseline, we use the per-example outcome averaged over 100 random routing repetitions. We report percentile-based 95\% confidence intervals. These intervals are nominal and are not adjusted for multiple comparisons.

\section{Results}
\label{sec:results}

This section provides a comprehensive analysis of confidence-based small-to-large routing across tasks, model families, and languages. We examine routing headroom, accuracy and efficiency trade-offs, confidence reliability, and the distribution of large-model access across languages.

\begin{figure*}[t]
    \centering
    \includegraphics[height=8cm, width=15cm]{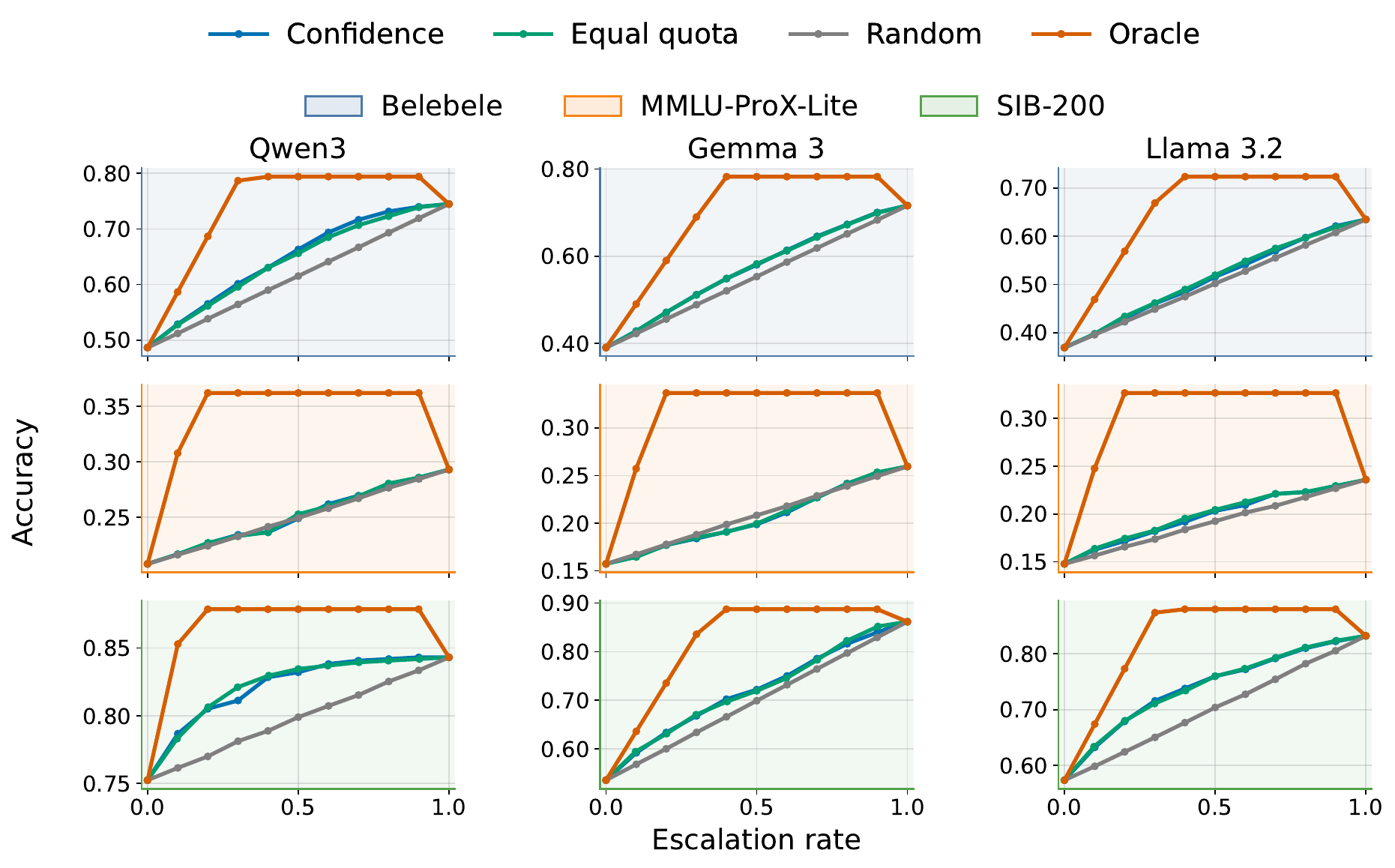}
    \caption{ Accuracy as a function of the escalation budget, macro-averaged over four languages. Confidence routes the lowest-margin examples, while equal quota applies confidence ranking separately within each language. Random routing provides the cost-matched baseline. The oracle uses the observed small-to-large correctness transition and is therefore an unattainable upper bound. Confidence provides consistent gains on Belebele and SIB-200 but remains close to random routing on MMLU-ProX-Lite. }
    \label{fig:routing_curves}
\end{figure*}

\subsection{Large Models Provide Task-Dependent Routing Headroom}
\label{sec:model_accuracy_results}
Table~\ref{tab:model_accuracy_auc} reports the macro-average accuracy over the four evaluation languages. The large model outperformed its corresponding small model in every dataset, family combination. However, the magnitude of this improvement varied considerably across tasks. On Belebele, replacing the small model with the large model improved accuracy by 25.81--32.64 percentage points. The corresponding improvement on SIB-200 ranged from 9.07 points for Qwen3 to 32.60 points for Gemma~3.

The gains were substantially smaller on MMLU-ProX-Lite, ranging from 8.51 to 10.22 points. Moreover, even the large models obtained only 23.59--29.29\% accuracy on this benchmark. Thus, MMLU-ProX-Lite presents a different routing regime: escalation offers measurable headroom, but the large model also fails on most examples.

\begin{table}[htbp]
\centering
\scriptsize
\setlength{\tabcolsep}{5.4pt}
\begin{tabular}{llrrrr}
\toprule
Dataset & Family & Small & Large & $\Delta$ & AUC \\
\midrule
\multirow{3}{*}{Belebele}
 & Qwen3     & 48.67 & 74.47 & +25.81 & 0.673 \\
 & Gemma 3   & 39.00 & 71.64 & +32.64 & 0.589 \\
 & Llama 3.2 & 36.89 & 63.47 & +26.58 & 0.574 \\
\midrule
\multirow{3}{*}{MMLU-ProX-Lite}
 & Qwen3     & 20.78 & 29.29 &  +8.51 & 0.499 \\
 & Gemma 3   & 15.73 & 25.95 & +10.22 & 0.488 \\
 & Llama 3.2 & 14.78 & 23.59 &  +8.81 & 0.544 \\
\midrule
\multirow{3}{*}{SIB-200}
 & Qwen3     & 75.25 & 84.31 &  +9.07 & 0.801 \\
 & Gemma 3   & 53.55 & 86.15 & +32.60 & 0.609 \\
 & Llama 3.2 & 57.35 & 83.21 & +25.86 & 0.704 \\
\bottomrule
\end{tabular}
\caption{Macro-average accuracy across English, Bangla, Hindi, and Urdu. $\Delta$ is the large-minus-small accuracy difference in percentage points. AUC measures how well low small-model confidence identifies errors corrected by the large model.}
\label{tab:model_accuracy_auc}
\end{table}

\subsection{Confidence Routing Is Strongly Task-Dependent}
\label{sec:routing_results}

Figure~\ref{fig:routing_curves} presents the complete accuracy, escalation trade-off. Confidence routing consistently outperformed random escalation on Belebele and SIB-200, whereas its advantage was small or absent on MMLU-ProX-Lite. The contrast remains visible across model families and escalation budgets, indicating that routing effectiveness is determined primarily by the task rather than by a single model family.

At a 50\% escalation budget, confidence routing significantly outperformed random routing for all three model families on Belebele and SIB-200 (Table~\ref{tab:routing50}). The paired-bootstrap improvement was 1.40--4.67 points on Belebele and 2.54--5.63 points on SIB-200. In contrast, confidence routing did not significantly outperform random routing for Qwen3 or Gemma~3 on MMLU-ProX-Lite. Only Llama~3.2 showed a small improvement of 1.14 points (95\% CI: 0.11--2.20).

\begin{table*}[t]
\centering
\small
\setlength{\tabcolsep}{5pt}
\begin{tabular}{llrrrr}
\toprule
Dataset & Family &
Confidence acc. &
$\Delta$ vs.\ random [95\% CI] &
$\Delta$ vs.\ EQ [95\% CI] &
Oracle acc. \\
\midrule
\multirow{3}{*}{Belebele}
 & Qwen3
 & 66.31
 & \cellcolor{sigpos}\textbf{$+4.67^{*}$ [3.53, 5.82]}
 & \cellcolor{sigpos}\textbf{$+0.69^{*}$ [0.06, 1.31]}
 & 79.39 \\
 & Gemma 3
 & 58.03
 & \cellcolor{sigpos}\textbf{$+2.73^{*}$ [1.46, 3.98]}
 & $-0.19$ [$-0.56$, 0.14]
 & 78.28 \\
 & Llama 3.2
 & 51.58
 & \cellcolor{sigpos}\textbf{$+1.40^{*}$ [0.34, 2.50]}
 & $-0.33$ [$-1.03$, 0.36]
 & 72.36 \\
\midrule
\multirow{3}{*}{MMLU-ProX-Lite}
 & Qwen3
 & 24.89
 & $-0.15$ [$-1.07$, 0.76]
 & \cellcolor{signeg}\textbf{$-0.38^{*}$ [$-0.76$, $-0.04$]}
 & 36.21 \\
 & Gemma 3
 & 19.87
 & $-0.99$ [$-2.11$, 0.13]
 & $-0.08$ [$-0.34$, 0.19]
 & 33.66 \\
 & Llama 3.2
 & 20.33
 & \cellcolor{sigpos}\textbf{$+1.14^{*}$ [0.11, 2.20]}
 & $-0.11$ [$-0.53$, 0.30]
 & 32.67 \\
\midrule
\multirow{3}{*}{SIB-200}
 & Qwen3
 & 83.21
 & \cellcolor{sigpos}\textbf{$+3.52^{*}$ [1.94, 5.16]}
 & $-0.25$ [$-0.86$, 0.49]
 & 87.87 \\
 & Gemma 3
 & 72.18
 & \cellcolor{sigpos}\textbf{$+2.54^{*}$ [0.14, 4.96]}
 & $+0.25$ [$-0.25$, 0.74]
 & 88.73 \\
 & Llama 3.2
 & 75.98
 & \cellcolor{sigpos}\textbf{$+5.63^{*}$ [3.24, 8.02]}
 & $0.00$ [$-0.61$, 0.61]
 & 87.99 \\
\bottomrule
\end{tabular}
\caption{Routing results at a 50\% escalation budget. Accuracy and differences are reported in percentage points. Differences and 95\% confidence intervals are obtained from 5,000 paired bootstrap samples over parallel example IDs. Green and red cells indicate significant positive and negative differences, respectively. $^{*}$ indicates that the interval excludes zero. EQ denotes equal-quota confidence routing. }
\label{tab:routing50}
\end{table*}

The oracle results show that the failure on MMLU-ProX-Lite cannot be explained solely by a lack of correctable examples. At 50\% escalation, oracle accuracy reached 32.67--36.21\%, compared with 19.87--24.89\% under confidence routing. Thus, substantial routing headroom remains, but the small model's confidence does not reliably identify the examples on which escalation is beneficial.

The measured system results confirm that escalation rate generally translates into realized savings, although the relationship is model dependent. At 25\% escalation, confidence routing reduces wall time by 27.5--58.2\% and energy by 26.3--62.6\% relative to large-only inference. At 50\%, however, savings fall to 0.9--32.3\% for wall time and $-0.5$--36.8\% for energy. Confidence routing can therefore approach large-only cost when it selects disproportionately long inputs. Complete throughput, energy, GPU-hour, and tail-latency results appear in Appendices~\ref{app:system_efficiency_results} and~\ref{app:tail_latency}.

\subsection{Why Confidence Fails on Multilingual Reasoning}
\label{sec:confidence_failure}

\begin{figure}[t]
    \centering
    \includegraphics[
        width=\columnwidth
    ]{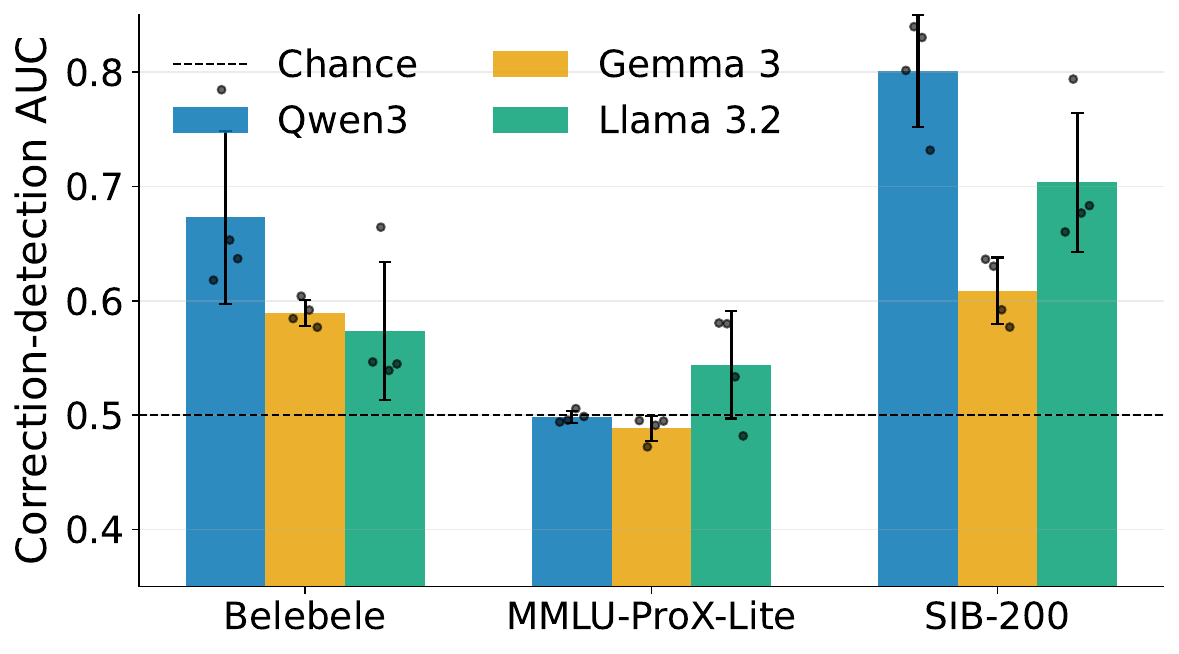}
\caption{Correction-detection AUC using low small-model confidence as the routing score. Bars show means across languages, error bars show standard deviations across languages, points show individual languages, and the dashed line denotes chance performance.}
    \label{fig:correction_auc}
\end{figure}

To examine the source of this task dependence, we treat a small-model error that is corrected by the large model as a positive routing event. Figure~\ref{fig:correction_auc} reports the AUC obtained when small-model confidence is used to detect these events. Averaged over model families, the correction-detection AUC was 0.612 on Belebele and 0.705 on SIB-200. In contrast, the average AUC on MMLU-ProX-Lite was only 0.510, close to chance.

This failure was consistent across families. MMLU-ProX-Lite AUC was 0.488 for Gemma~3, 0.499 for Qwen3, and 0.544 for Llama~3.2. Therefore, although the large models corrected 15.43--17.93\% of all MMLU-ProX-Lite examples, these corrections were not concentrated among the examples assigned low confidence by the small model.

The confidence-bin analysis provides further evidence of confident failure. In the highest-confidence quintile of MMLU-ProX-Lite, the small-model error rate remained 63.45\% for Qwen3, 76.89\% for Gemma~3, and 86.17\% for Llama~3.2. Furthermore, only 15.59--21.89\% of these high-confidence errors were corrected by the corresponding large model. Confidence therefore captures neither small-model correctness nor the conditional benefit of escalation reliably on this task.

Small and large models also frequently failed on the same reasoning questions. Both models were incorrect on 63.79--67.33\% of MMLU-ProX-Lite examples, compared with 20.61--27.64\% on Belebele and 11.27--12.13\% on SIB-200. This high co-failure rate explains the low absolute MMLU-ProX-Lite accuracy, while the near-chance AUC explains why the remaining correctable errors are difficult to route.

\subsection{Language Allocation and Routing Disparity}
\label{sec:language_disparity_results}

Global confidence ranking did not distribute escalations uniformly across languages. At a nominal 50\% budget, the difference between the most- and least-escalated language ranged from 3.43 to 30.33 percentage points across the nine dataset, family settings (Table~\ref{tab:language_allocation}). The largest gaps occurred for Qwen3 on Belebele (30.33 points) and SIB-200 (25.49 points), and for Llama~3.2 on Belebele (25.78 points).

Equal-quota routing eliminates this allocation disparity by construction. Its aggregate accuracy was statistically indistinguishable from global confidence routing in seven of the nine settings in Table~\ref{tab:routing50}. The two exceptions were Qwen3: global confidence was 0.69 points better on Belebele, whereas equal quota was 0.38 points better on MMLU-ProX-Lite. However, equal allocation did not consistently reduce the final language accuracy gap. Resource parity and outcome parity should therefore be reported as separate properties of a routing policy.

\begin{table}[htbp]
\centering
\scriptsize
\setlength{\tabcolsep}{4pt}
\begin{tabular}{llrrr}
\toprule
Dataset & Family &
Esc.\ gap &
\multicolumn{2}{c}{Worst-language accuracy} \\
\cmidrule(lr){4-5}
 & & Confidence & Confidence & Equal quota \\
\midrule
\multirow{3}{*}{Belebele}
 & Qwen3     & \cellcolor{gaphigh}\textbf{30.33} & 58.11 & 56.33 \\
 & Gemma 3   & \cellcolor{gapmid}8.67             & 50.44 & 50.22 \\
 & Llama 3.2 & \cellcolor{gaphigh}25.78            & 43.78 & 42.67 \\
\midrule
\multirow{3}{*}{MMLU-ProX-Lite}
 & Qwen3     & \cellcolor{gapmid}11.25 & 21.28 & 21.43 \\
 & Gemma 3   & \cellcolor{gaplow}4.56  & 16.87 & 16.87 \\
 & Llama 3.2 & \cellcolor{gapmid}10.49 & 18.09 & 17.78 \\
\midrule
\multirow{3}{*}{SIB-200}
 & Qwen3     & \cellcolor{gaphigh}25.49 & 80.39 & 79.41 \\
 & Gemma 3   & \cellcolor{gaplow}3.43   & 66.67 & 66.67 \\
 & Llama 3.2 & \cellcolor{gaplow}3.92   & 68.63 & 68.63 \\
\bottomrule
\end{tabular}
\caption{Language allocation at 50\% escalation. Esc.\ gap is the highest-minus-lowest language escalation rate under global confidence routing, with darker shading indicating greater disparity. The final columns report worst-language accuracy under global confidence and equal-quota routing. Equal quota has zero escalation gap by construction.}
\label{tab:language_allocation}
\end{table}

Table~\ref{tab:who_pays} makes the language-level allocation consequences explicit. Although the nominal global budget is 50\%, English received an average escalation rate of only 43.98\% and was the least-escalated language in five of the nine settings. Equal-quota routing increased its mean routed accuracy from 61.11\% to 62.46\%, corresponding to a 1.35-point improvement. In contrast, Bangla, Hindi, and Urdu received average escalation rates at or above the nominal budget, and global routing provided small mean accuracy advantages of 0.26--0.56 points over equal quota.
\begin{table}[htbp]
\centering
\scriptsize
\setlength{\tabcolsep}{7.2pt}
\begin{tabular}{lccrrr}
\toprule
& \multicolumn{2}{c}{Allocation}
& \multicolumn{3}{c}{Routed accuracy (\%)} \\
\cmidrule(lr){2-3}
\cmidrule(lr){4-6}
Language
& \shortstack{Least\\(of 9)}
& \shortstack{Mean esc.\\rate (\%)}
& Global
& EQ
& $\Delta$ \\
\midrule
English & 5/9 & 43.98 & 61.11 & 62.46 & $-1.35$ \\
Bangla  & 3/9 & 53.04 & 49.06 & 48.50 & $+0.56$ \\
Hindi   & 0/9 & 50.85 & 51.86 & 51.60 & $+0.26$ \\
Urdu    & 1/9 & 52.14 & 47.91 & 47.56 & $+0.36$ \\
\bottomrule
\end{tabular}
\caption{\textbf{Who pays for efficient inference?} Language-level allocation and routed accuracy at 50\% escalation, averaged across nine dataset and model family settings. $\Delta$ is global minus equal-quota accuracy.}
\label{tab:who_pays}
\end{table}

These results provide a more precise answer to \texttt{who pays for efficient inference}. In aggregate, English bears the clearest allocation and accuracy cost under global confidence routing. However, this pattern is not universal: Bangla received the fewest escalations in three settings and Urdu in one, while Hindi was never the least-escalated language. The burden therefore shifts across dataset and model family settings rather than following a fixed distinction between high-resource and low-resource languages. Moreover, the comparatively small differences between global and equal-quota accuracy show that removing allocation disparity generally requires little aggregate language-level performance sacrifice.

\section{Conclusion}
\label{sec:conclusion}

This research shows that small-to-large escalation provides substantial but task-dependent accuracy and efficiency gains. Confidence identified beneficial escalations for reading comprehension and topic classification, but was nearly uninformative for multilingual reasoning, where small models frequently remained confident while incorrect. At 25\% escalation, confidence routing reduced measured wall time by 27.5--58.2\% and energy by 26.3--62.6\% relative to large-only inference, although these benefits diminished substantially at 50\% escalation. Global confidence routing also distributed large-model access unevenly, producing language escalation gaps of up to 30.33 percentage points, whereas equal-quota routing removed this disparity with little aggregate accuracy cost. These results answer the question posed by the title, \textbf{who pays for efficient inference?} No single language consistently bears the allocation cost. Instead, the burden depends on the interaction among task, model family, language, and confidence distribution rather than a fixed division between high-resource and low-resource languages. Equal-quota routing guarantees equal access to the larger model, but does not necessarily eliminate differences in final language performance, confirming that allocation parity and outcome parity are distinct properties. Overall, confidence should not be treated as a task-independent or language-neutral routing signal. Future work should estimate escalation benefit and incorporate multilingual allocation into routing decisions.

\section{Limitations}
\label{sec:limitations}

Our evaluation is limited to three open model families between 1B and 4B parameters. The findings may not generalize to substantially larger or proprietary models, mixture-of-experts architectures, or model pairs with different capability gaps. The unequal parameter increases within each family may also affect the available routing headroom.

We study four languages and three multiple-choice benchmarks covering reading comprehension, knowledge and reasoning, and topic classification. These settings do not represent broader linguistic diversity or free-form tasks such as dialogue, summarization, and translation. Moreover, our confidence measure is the probability margin between the two highest-scoring answer labels. Other uncertainty signals or confidence derived from unrestricted generation may behave differently.

We use escalation rate as the primary policy-level cost measure and supplement it with system measurements using sequential, batch-size-one inference on two NVIDIA Tesla T4 GPUs. The measured efficiency may differ under batching, concurrent serving, queueing, communication overhead, model loading, or other hardware. Energy includes both allocated GPUs, while GPU-hours are reported instead of vendor-dependent monetary cost. Gemma~3 4B uses FP32, so absolute efficiency should not be compared directly across families.

Finally, our analysis uses offline paired predictions from both models. This enables controlled comparisons but does not reproduce every aspect of live deployment. The oracle uses ground-truth correctness only as a diagnostic upper bound and is not deployable.

\bibliography{custom}

\appendix

\section*{Appendix}

\section{Additional Dataset Details}
\label{app:data}

\subsection{Dataset Descriptions}

\paragraph{Belebele} is a parallel multilingual reading comprehension benchmark \citep{bandarkar-etal-2024-belebele}. Each example contains a passage, a question, and four candidate answers. We use the complete test split, which provides 900 examples per language. Because equivalent content is available across the selected languages, the benchmark supports controlled comparisons of model performance and routing behavior across languages.

\paragraph{MMLU-ProX-Lite} is a multilingual knowledge and reasoning benchmark derived from multiple-choice academic questions \citep{xuan-etal-2025-mmlu}. We use the Lite configuration and combine its validation and test partitions, producing 658 examples per language. This combination is appropriate for our primary routing policies because they neither learn parameters nor select thresholds using ground-truth labels. Oracle routing uses labels only as a diagnostic upper bound and is never treated as a deployable method.

\paragraph{SIB-200} is a multilingual topic classification benchmark covering a broad range of languages and scripts \citep{adelani-etal-2024-sib}. We use the complete test split, which contains 204 examples per language. We formulate each example as a constrained choice among symbolic answer labels corresponding to the available topic categories. This formulation allows SIB-200 to follow the same deterministic scoring and routing procedure used for the other benchmarks.

\subsection{Dataset Preparation and Pairing}
\label{app:preprocess}

Each dataset was converted to a common JSON Lines representation containing the dataset and task identifiers, example ID, language code, question or input text, optional context, answer choices, symbolic choice labels, and correct label. Belebele uses its test split, MMLU-ProX-Lite combines the validation and test splits, and SIB-200 uses its test split. The expected number of examples was checked during preprocessing: 900 per language for Belebele, 658 for MMLU-ProX-Lite, and 204 for SIB-200. The same processed examples and symbolic labels were supplied to both model sizes within a family.

Small and large predictions were joined using dataset, family, language, and example ID. Routing analyses were performed only after verifying a one-to-one match between the two model outputs. This produced 21,144 paired small-large comparisons, corresponding to 42,288 individual predictions. Parallel language versions retain a shared question or example identity where available, enabling the statistical procedure to preserve cross-language dependence.

\section{Implementation and Reproducibility Details}
\label{app:implementation}

This section provides implementation details omitted from the main paper for space. The evaluation pipeline was designed to preserve exact pairing between the small and large models, record the provenance of every prediction, and permit interrupted GPU runs to resume without changing the evaluated examples.

\subsection{Reproducibility and Experiment Management}
\label{app:reproducibility}

All experiments use a common configuration-driven pipeline that separates model, dataset, inference, and routing settings from the evaluation logic. The same scoring procedure is applied across model families, sizes, datasets, and languages. Model repository revisions are resolved and recorded before evaluation, while the Git commit and software environment are preserved with the results. Table~\ref{tab:software_environment} summarizes the recorded experimental environment.

\begin{table}[htbp]
    \centering
    \scriptsize
    \setlength{\tabcolsep}{22pt}
    \begin{tabular}{ll}
        \toprule
        Component & Version or platform \\
        \midrule
        PyTorch & 2.10.0+cu128 \\
        Transformers & 4.57.6 \\
        Accelerate & 1.13.0 \\
        Datasets & 4.8.5 \\
        Hugging Face Hub & 0.36.2 \\
        CUDA runtime & 12.8 \\
        Environment & Kaggle \\
        Hardware & NVIDIA Tesla T4 GPUs \\
        \bottomrule
    \end{tabular}
    \caption{Software and hardware environment used for model evaluation.}
    \label{tab:software_environment}
\end{table}

\subsection{Prompt Construction and Answer Scoring}
\label{app:prompt_scoring}

We use task-specific prompts with a common constrained-answer instruction. For Belebele, the prompt contains the passage, question, and four answer choices and asks the model to return one symbolic label. For MMLU-ProX-Lite, it contains the question and available choices without a passage. For SIB-200, the input text and topic names are presented as a constrained topic-classification problem. The model's native instruction template is then applied with generation-prompt tokens enabled. Qwen3 thinking mode is disabled so that all families are evaluated through the same direct-answer interface.

We score each example from the next-token logits rather than generating free-form text. Every symbolic answer label is first verified to correspond to exactly one tokenizer token, using the plain label or its space-prefixed form when necessary. Let $z_{ij}$ denote the next-token logit assigned to choice $j$ for example $i$. We normalize only across the valid answer labels,
\begin{equation}
    p_{ij}=\frac{\exp(z_{ij})}{\sum_{k\in\mathcal{Y}_i}\exp(z_{ik})},
\end{equation}
where $\mathcal{Y}_i$ is the set of candidate labels. The predicted label is the choice with maximum probability. The confidence score used for routing is the margin between the two highest normalized choice probabilities,
\begin{equation}
    m_i=p_{i(1)}-p_{i(2)}.
\end{equation}
This construction compares confidence only among valid task choices and avoids variation caused by unconstrained answer parsing.

\subsection{Model Inference}
\label{app:model_inference}

Models and tokenizers were loaded from Hugging Face using the exact repository revision returned at evaluation time, and the revision hash was written into every result row. All models were placed in evaluation mode and scored under \texttt{torch.inference\_mode()}. Inputs were truncated to at most 4,096 tokens, inference used a batch size of one, and key-value caching was disabled. No sampling or text generation was performed. On the NVIDIA T4 environment, models ordinarily used float16, while hardware supporting bfloat16 could select that datatype automatically. Each output row additionally records the model identifier, revision, datatype, seed, software environment, input-token count, and measured latency.

Gemma~3 4B required a precision-specific correction. Its float16 forward pass produced nonfinite vocabulary logits on the affected MMLU-ProX-Lite and SIB-200 evaluations. We therefore reran those configurations in float32 across two GPUs, with the language-model layers distributed by the automatic device map. Because only the final position is required, the corrected implementation requests one output-logit position for this model. It also explicitly tests that every selected answer logit is finite and stops the evaluation if this condition is violated. The corrected files were accepted only when they contained the expected number of unique examples, recorded float32 as the datatype, and contained finite confidence values. All analyses use these corrected outputs.

\begin{figure*}[t]
    \centering
    \includegraphics[width=\textwidth]{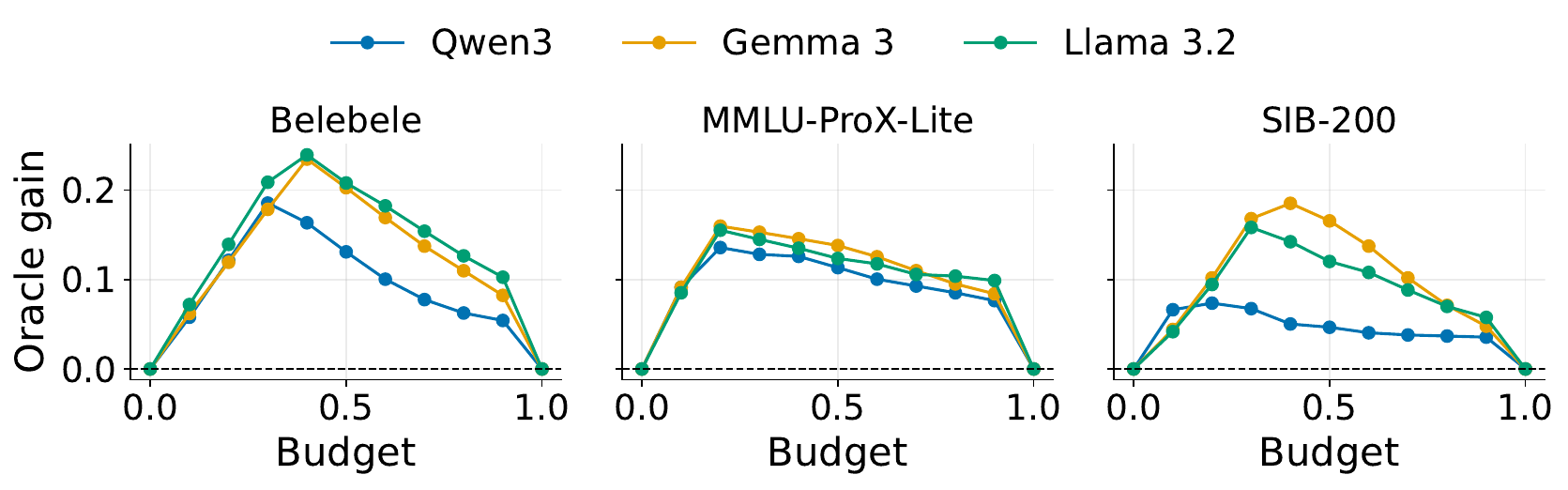}
    \caption{Accuracy difference between oracle and confidence routing. Positive gaps at intermediate budgets show that confidence leaves useful correction opportunities unidentified.}
    \label{fig:app_oracle_headroom}
\end{figure*}

\begin{figure*}[t]
    \centering
    \includegraphics[width=\textwidth]{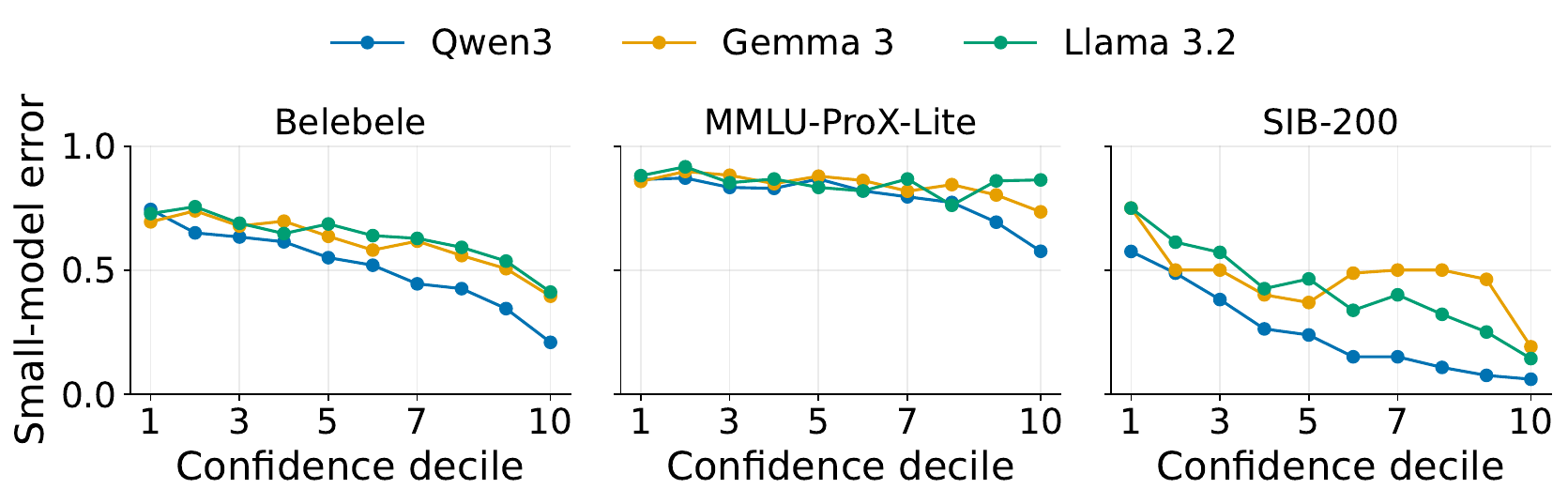}
    \caption{Small-model error rate by confidence decile, averaged across languages. Error falls clearly with confidence on Belebele and SIB-200, but remains high and comparatively flat on MMLU-ProX-Lite.}
    \label{fig:app_error_deciles}
\end{figure*}

\subsection{Routing Implementation}
\label{app:routing_implementation}

For every dataset and model family, the four languages are concatenated before global routing. At budget $b$, the router escalates $\mathrm{round}(bN)$ of the $N$ examples. Global confidence routing selects the examples with the smallest $m_i$. Equal-quota routing applies the same ranking independently within each language and escalates approximately the same fraction from every language. Random routing samples the budgeted number of examples without replacement and averages the per-example outcomes over 100 seeded repetitions. Oracle routing ranks examples by the realized correctness gain $l_i-s_i$ and serves only as a diagnostic upper bound. 
\begin{table}[htbp]
    \centering
    \scriptsize
    \setlength{\tabcolsep}{20pt}
    \begin{tabular}{ll}
        \toprule
        Setting & Value \\
        \midrule
        Random seed & 42 \\
        Maximum input length & 4,096 tokens \\
        Evaluation batch size & 1 \\
        Checkpoint interval & 25 examples \\
        Routing budgets & 0.0 to 1.0 by 0.1 \\
        Random-routing repetitions & 100 \\
        Bootstrap repetitions & 5,000 \\
        \bottomrule
    \end{tabular}
    \caption{Principal evaluation and analysis settings.}
    \label{tab:app_reproducibility}
\end{table}

Routing curves are evaluated from zero to full escalation in increments of 0.1. For an escalated example, routed correctness is taken from the large model; otherwise it is taken from the small model. Stable sorting is used for confidence and oracle rankings so that ties have deterministic behavior. The principal fixed-budget comparisons use 25\% and 50\% escalation. Table~\ref{tab:app_reproducibility} summarizes the main implementation settings.

\subsection{Uncertainty Estimation}
\label{app:uncertainty_details}

The final comparisons use 5,000 paired clustered-bootstrap repetitions. Within each dataset and family, resampling is performed over the shared example or question IDs, and all corresponding language versions are retained in the same sampled cluster. The routed outcomes of the compared policies therefore remain paired, while dependence among parallel multilingual examples is preserved. For the random baseline, the outcome associated with each example is first averaged over the 100 random-routing repetitions. We report the observed paired accuracy difference and the 2.5th and 97.5th percentiles of the bootstrap distribution. The intervals are nominal and are not adjusted for multiple comparisons.

\begin{table*}[t]
    \centering
    \scriptsize
    \setlength{\tabcolsep}{8.5pt}
    \begin{tabular}{llcccccccccc}
        \toprule
        & & \multicolumn{2}{c}{English} & \multicolumn{2}{c}{Bangla} & \multicolumn{2}{c}{Hindi} & \multicolumn{2}{c}{Urdu} & \multicolumn{2}{c}{Mean} \\
        \cmidrule(lr){3-4}\cmidrule(lr){5-6}\cmidrule(lr){7-8}\cmidrule(lr){9-10}\cmidrule(lr){11-12}
        Dataset & Family & Small & Large & Small & Large & Small & Large & Small & Large & Small & Large \\
        \midrule
        Belebele & Qwen3 & 72.22 & 88.33 & 38.78 & 69.22 & 43.00 & 70.44 & 40.67 & 69.89 & 48.67 & 74.47 \\
        & Gemma 3 & 50.22 & 87.00 & 36.89 & 67.56 & 36.11 & 68.44 & 32.78 & 63.56 & 39.00 & 71.64 \\
        & Llama 3.2 & 48.67 & 84.56 & 30.11 & 53.44 & 36.89 & 59.78 & 31.89 & 56.11 & 36.89 & 63.47 \\
        \midrule
        MMLU-ProX-Lite & Qwen3 & 29.94 & 37.99 & 16.72 & 25.84 & 17.02 & 25.99 & 19.45 & 27.36 & 20.78 & 29.29 \\
        & Gemma 3 & 17.63 & 31.00 & 13.83 & 22.04 & 15.81 & 26.29 & 15.65 & 24.47 & 15.73 & 25.95 \\
        & Llama 3.2 & 13.98 & 32.83 & 14.44 & 19.15 & 16.72 & 21.43 & 13.98 & 20.97 & 14.78 & 23.59 \\
        \midrule
        SIB-200 & Qwen3 & 78.43 & 86.76 & 73.04 & 83.33 & 76.47 & 85.29 & 73.04 & 81.86 & 75.25 & 84.31 \\
        & Gemma 3 & 59.31 & 88.24 & 54.41 & 86.27 & 55.88 & 85.29 & 44.61 & 84.80 & 53.55 & 86.15 \\
        & Llama 3.2 & 70.59 & 88.73 & 50.98 & 79.41 & 59.31 & 84.80 & 48.53 & 79.90 & 57.35 & 83.21 \\
        \bottomrule
    \end{tabular}
    \caption{Accuracy (\%) of the small and large models by dataset, family, and language. Mean denotes the unweighted average across the four languages.}
    \label{tab:app_model_accuracy_language}
\end{table*}

\begin{table*}[t]
\centering
\scriptsize
\setlength{\tabcolsep}{13pt}
\begin{tabular}{lllcc}
\toprule
Dataset & Family & Comparison & 25\% escalation & 50\% escalation \\
\midrule
\multirow{6}{*}{Belebele}
& \multirow{2}{*}{Qwen3}
& Confidence $-$ Random
& \cellcolor{sigpos}\textbf{$+3.29\ [ +2.27, +4.34 ]$}
& \cellcolor{sigpos}\textbf{$+4.67\ [ +3.53, +5.82 ]$} \\
&
& Global confidence $-$ Equal quota
& \cellcolor{sigpos}\textbf{$+0.58\ [ +0.06, +1.08 ]$}
& \cellcolor{sigpos}\textbf{$+0.69\ [ +0.06, +1.31 ]$} \\

& \multirow{2}{*}{Gemma 3}
& Confidence $-$ Random
& \cellcolor{sigpos}\textbf{$+1.89\ [ +0.83, +2.95 ]$}
& \cellcolor{sigpos}\textbf{$+2.73\ [ +1.46, +3.98 ]$} \\
&
& Global confidence $-$ Equal quota
& $-0.31\ [ -0.64, +0.03 ]$
& $-0.19\ [ -0.56, +0.14 ]$ \\

& \multirow{2}{*}{Llama 3.2}
& Confidence $-$ Random
& $+0.95\ [ -0.03, +1.89 ]$
& \cellcolor{sigpos}\textbf{$+1.40\ [ +0.34, +2.50 ]$} \\
&
& Global confidence $-$ Equal quota
& $-0.33\ [ -0.83, +0.22 ]$
& $-0.33\ [ -1.03, +0.36 ]$ \\
\midrule

\multirow{6}{*}{MMLU-ProX-Lite}
& \multirow{2}{*}{Qwen3}
& Confidence $-$ Random
& $+0.05\ [ -0.76, +0.82 ]$
& $-0.15\ [ -1.07, +0.76 ]$ \\
&
& Global confidence $-$ Equal quota
& $+0.11\ [ -0.15, +0.38 ]$
& \cellcolor{signeg}\textbf{$-0.38\ [ -0.76, -0.04 ]$} \\

& \multirow{2}{*}{Gemma 3}
& Confidence $-$ Random
& $+0.11\ [ -0.81, +1.10 ]$
& $-0.99\ [ -2.11, +0.13 ]$ \\
&
& Global confidence $-$ Equal quota
& $+0.04\ [ +0.00, +0.11 ]$
& $-0.08\ [ -0.34, +0.19 ]$ \\

& \multirow{2}{*}{Llama 3.2}
& Confidence $-$ Random
& $+0.73\ [ -0.18, +1.64 ]$
& \cellcolor{sigpos}\textbf{$+1.14\ [ +0.11, +2.20 ]$} \\
&
& Global confidence $-$ Equal quota
& $+0.11\ [ -0.19, +0.42 ]$
& $-0.11\ [ -0.53, +0.30 ]$ \\
\midrule

\multirow{6}{*}{SIB-200}
& \multirow{2}{*}{Qwen3}
& Confidence $-$ Random
& \cellcolor{sigpos}\textbf{$+3.75\ [ +1.86, +5.64 ]$}
& \cellcolor{sigpos}\textbf{$+3.52\ [ +1.94, +5.16 ]$} \\
&
& Global confidence $-$ Equal quota
& $-0.25\ [ -1.10, +0.61 ]$
& $-0.25\ [ -0.86, +0.49 ]$ \\

& \multirow{2}{*}{Gemma 3}
& Confidence $-$ Random
& \cellcolor{sigpos}\textbf{$+3.94\ [ +1.77, +6.19 ]$}
& \cellcolor{sigpos}\textbf{$+2.54\ [ +0.14, +4.96 ]$} \\
&
& Global confidence $-$ Equal quota
& $+0.61\ [ -0.12, +1.35 ]$
& $+0.25\ [ -0.25, +0.74 ]$ \\

& \multirow{2}{*}{Llama 3.2}
& Confidence $-$ Random
& \cellcolor{sigpos}\textbf{$+6.27\ [ +3.90, +8.66 ]$}
& \cellcolor{sigpos}\textbf{$+5.63\ [ +3.24, +8.02 ]$} \\
&
& Global confidence $-$ Equal quota
& $+0.25\ [ -0.37, +0.86 ]$
& $+0.00\ [ -0.61, +0.61 ]$ \\
\bottomrule
\end{tabular}
\caption{Paired accuracy differences in percentage points between confidence routing and the random and equal-quota baselines, with 95\% clustered-bootstrap confidence intervals. Positive values favor the first policy named in each comparison. Green and red cells indicate statistically significant positive and negative differences, respectively.}
\label{tab:app_bootstrap_comparisons}
\end{table*}

\subsection{Checkpointing and Result Validation}
\label{app:validation}

Evaluation results are checkpointed every 25 newly scored examples. When a run resumes, completed example IDs are loaded from the existing CSV and skipped. Before final saving, duplicate example IDs are removed while retaining the most recent completed record. This design allowed long model and language runs to be divided across Kaggle sessions without changing their predictions or example ordering.

The final analysis begins from a frozen canonical result archive rather than searching across earlier outputs. It checks the expected dataset size, uniqueness of example IDs, availability of both model sizes, agreement of task metadata, and finiteness of confidence values. Smoke-test files carrying limit suffixes and invalid Gemma~3 4B backups are excluded. Intermediate tables, figures, and statistical comparisons are regenerated deterministically from the paired prediction table, preserving a direct path from each reported aggregate to its underlying examples.

\subsection{System Efficiency Setup}
\label{app:system_efficiency}

We measure system efficiency using the same deterministic sample of 1,200 inputs for every model family. The sample contains 100 examples from each of the twelve dataset and language combinations, covering three datasets and four languages. All measurements use two NVIDIA Tesla T4 GPUs with batch size one. Each model processes five warm-up examples before measurement, and every condition is repeated three times. CUDA execution is synchronized around each timed inference operation.

Small-only inference processes all inputs with the smaller model, while large-only inference processes all inputs with the larger model. Routing is implemented as a sequential cascade in which the smaller model first evaluates all inputs and the larger model subsequently evaluates only the selected subset. At 25\% and 50\% escalation, the larger model therefore processes 300 and 600 examples, respectively. Model loading, checkpoint download, and dataset preparation are excluded because they represent one-time rather than per-query costs.

\begin{table*}[t]
\centering
\scriptsize
\setlength{\tabcolsep}{12pt}
\begin{tabular}{llcccccccc}
\toprule
& & \multicolumn{2}{c}{English} & \multicolumn{2}{c}{Bangla}
& \multicolumn{2}{c}{Hindi} & \multicolumn{2}{c}{Urdu} \\
\cmidrule(lr){3-4}\cmidrule(lr){5-6}
\cmidrule(lr){7-8}\cmidrule(lr){9-10}
Dataset & Family & Esc. & Acc. & Esc. & Acc. & Esc. & Acc. & Esc. & Acc. \\
\midrule
Belebele & Qwen3
& \cellcolor{belowstrong}\textbf{28.33} & 81.56
& \cellcolor{abovestrong}\textbf{55.33} & 58.11
& \cellcolor{abovestrong}\textbf{58.67} & 64.33
& \cellcolor{abovestrong}\textbf{57.67} & 61.22 \\
& Gemma 3
& 49.67 & 71.11
& \cellcolor{belowlight}45.44 & 53.00
& \cellcolor{abovelight}54.11 & 57.56
& 50.78 & 50.44 \\
& Llama 3.2
& \cellcolor{belowstrong}\textbf{41.89} & 68.33
& \cellcolor{abovestrong}\textbf{67.67} & 47.56
& \cellcolor{belowstrong}\textbf{42.56} & 46.67
& \cellcolor{belowlight}47.89 & 43.78 \\
\midrule
MMLU-ProX-Lite & Qwen3
& \cellcolor{belowlight}46.96 & 33.89
& \cellcolor{abovestrong}\textbf{55.93} & 22.49
& \cellcolor{abovelight}52.43 & 21.28
& \cellcolor{belowstrong}\textbf{44.68} & 21.88 \\
& Gemma 3
& 48.33 & 23.10
& 48.63 & 16.87
& 50.15 & 20.36
& \cellcolor{abovelight}52.89 & 19.15 \\
& Llama 3.2
& \cellcolor{belowstrong}\textbf{44.83} & 24.47
& \cellcolor{abovestrong}\textbf{55.32} & 19.00
& \cellcolor{belowlight}47.26 & 19.76
& \cellcolor{abovelight}52.58 & 18.09 \\
\midrule
SIB-200 & Qwen3
& \cellcolor{belowstrong}\textbf{34.80} & 85.29
& \cellcolor{abovelight}52.45 & 81.37
& \cellcolor{abovelight}52.45 & 85.78
& \cellcolor{abovestrong}\textbf{60.29} & 80.39 \\
& Gemma 3
& 51.96 & 75.49
& 48.53 & 74.51
& 49.02 & 72.06
& 50.49 & 66.67 \\
& Llama 3.2
& 49.02 & 86.76
& 48.04 & 68.63
& 50.98 & 78.92
& 51.96 & 69.61 \\
\bottomrule
\end{tabular}
\caption{Language-level escalation rate and routed accuracy (\%) under global confidence routing at the nominal 50\% budget. Blue and orange cells indicate escalation below and above the nominal budget, respectively, with darker shading indicating larger deviations; unshaded cells are within two percentage points of 50\%.}
\label{tab:app_language_routing_50}
\end{table*}

Per-example latency includes prompt construction, tokenization, device transfer, and the model forward pass. For routing policy $\pi$, routed latency for example $i$ is
\begin{equation}
    L_i^{\pi}=L_i^{S}+z_i^{\pi}L_i^{L},
    \label{eq:routed_latency}
\end{equation}
where $L_i^{S}$ and $L_i^{L}$ denote small-model and large-model latency, and $z_i^{\pi}\in\{0,1\}$ indicates whether the example is escalated. Mean, median, and 95th-percentile latency are calculated from these per-example routed latencies.

System wall time additionally includes the measured routing-policy overhead:
\begin{equation}
    T_{\pi}=T_{S}+T_{L,\pi}+T_{\mathrm{rank},\pi},
    \label{eq:routing_wall_time}
\end{equation}
where $T_{S}$ is the time required to process all inputs with the smaller model, $T_{L,\pi}$ is the time required for the larger model to process the selected subset, and $T_{\mathrm{rank},\pi}$ is the policy-selection time. Throughput is calculated as
\begin{equation}
    Q_{\pi}=\frac{N}{T_{\pi}},
    \label{eq:routing_throughput}
\end{equation}
where $N=1{,}200$.

GPU power is sampled from both allocated GPUs every 50\,ms using NVML. Total energy is estimated through trapezoidal integration:

\begin{equation}
    E_{\mathrm{Wh}}
    =
    \frac{1}{3600}
    \sum_{j=1}^{m-1}
    \frac{P_j+P_{j+1}}{2}
    \left(t_{j+1}-t_j\right).
\end{equation}
where $P_j$ is the combined power of both GPUs at time $t_j$. Routed energy is the sum of the energy consumed during the small-model stage and the selected large-model stage. The negligible CPU-side ranking energy is not included. Because energy measurements cover both allocated GPUs, they include the idle power of any GPU not actively used during part of a stage.

Wall-time and energy savings are calculated relative to large-only inference within the same family:
\begin{equation}
    S_T(\pi)=1-\frac{T_{\pi}}{T_L},
    \qquad
    S_E(\pi)=1-\frac{E_{\pi}}{E_L},
    \label{eq:efficiency_savings}
\end{equation}
where $T_L$ and $E_L$ denote large-only wall time and energy. GPU-hours are calculated as
\begin{equation}
    H_{\pi}=\frac{2T_{\pi}}{3600},
    \label{eq:gpu_hours}
\end{equation}
because two GPUs are allocated throughout the benchmark. We do not assign a fixed monetary price because the experiments are conducted on Kaggle and cloud rates vary across providers and purchasing arrangements. Given a price of $p$ dollars per T4 GPU-hour, estimated accelerator cost is $C_{\pi}=pH_{\pi}$.

Qwen3 and Llama~3.2 are evaluated in FP16. Gemma~3 1B is also evaluated in FP16, whereas Gemma~3 4B uses FP32 to avoid the invalid logits observed under FP16. Absolute efficiency should therefore not be compared directly across model families. Within-family policy comparisons remain valid because they use the same models, precision settings, examples, and hardware.

\section{Additional Results}
\label{app:additional_results}

This appendix provides the complete numerical breakdowns and complementary analyses underlying the principal results. All values are computed from the canonical corrected result package. In particular, the Gemma~3 4B results for MMLU-ProX-Lite and SIB-200 use the corrected float32 evaluations. Results are reported for English (En), Bangla (Bn), Hindi (Hi), and Urdu (Ur). Unless stated otherwise, aggregate values are macro-averaged across the four languages.

To avoid duplicating the main paper, this section reports only detailed language-level results, additional budget comparisons, diagnostic analyses, and complete system-efficiency measurements.

\subsection{Model Accuracy by Language}
\label{app:model_accuracy}

Table~\ref{tab:app_model_accuracy_language} reports small and large model accuracy for every dataset, family, and language. The larger model improves accuracy in all 36 dataset, family, and language comparisons, although the magnitude of improvement varies substantially. The largest differences occur on Belebele and SIB-200, while both model sizes remain comparatively weak on MMLU-ProX-Lite.

\subsection{Fixed-Budget Statistical Comparisons}
\label{app:statistical_results}

Table~\ref{tab:app_bootstrap_comparisons} reports paired accuracy differences using 5,000 clustered-bootstrap repetitions. Relative to random routing, confidence routing produces clear gains on Belebele and SIB-200. The improvements are statistically significant for Qwen3 and Gemma~3 at both Belebele budgets, while Llama~3.2 reaches significance at the 50\% budget. All SIB-200 comparisons are statistically significant, with gains ranging from 3.75 to 6.27 percentage points at 25\% escalation and from 2.54 to 5.63 points at 50\%. In contrast, most MMLU-ProX-Lite confidence intervals include zero, indicating no reliable advantage over random routing. The only exception is Llama~3.2 at 50\% escalation, which improves accuracy by 1.14 points with a 95\% confidence interval of $[+0.11,+2.20]$.

Differences between global confidence and equal-quota confidence routing are substantially smaller. Qwen3 on Belebele is the only setting in which global confidence routing consistently achieves a statistically significant advantage at both budgets, improving accuracy by 0.58 and 0.69 points. Most remaining confidence intervals include zero, suggesting that enforcing equal language quotas generally has little effect on aggregate accuracy. On MMLU-ProX-Lite, equal-quota routing significantly outperforms global confidence routing for Qwen3 at 50\% escalation by 0.38 points, while the remaining differences are statistically uncertain. These findings show that equal-quota routing can remove cross-language allocation disparities without consistently sacrificing overall accuracy. Figure~\ref{fig:app_oracle_headroom} further shows that useful correction opportunities remain beyond those identified by confidence.

\subsection{Confidence Reliability and Error Analysis}
\label{app:confidence_reliability}

\begin{figure}[htbp]
    \centering
    \includegraphics[width=0.48\textwidth]{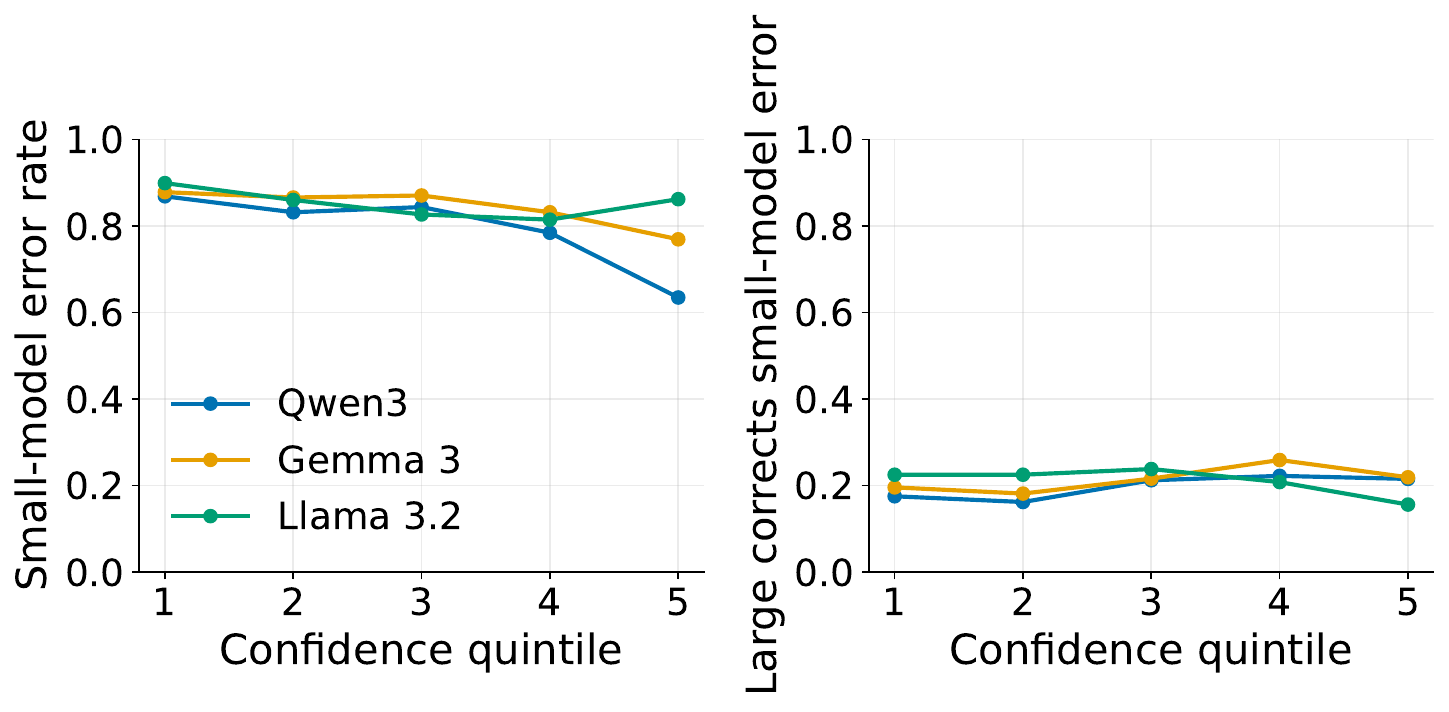}
    \caption{MMLU-ProX-Lite small-model error and correction rates across confidence quintiles. High-confidence examples retain high error rates, and correction opportunities are not concentrated consistently in the lowest-confidence quintiles.}
    \label{fig:app_mmlu_quintiles}
\end{figure}

The language-specific AUC values in Table~\ref{tab:app_auc_language} confirm that confidence reliability varies within as well as across tasks. MMLU-ProX-Lite remains near chance for most family and language combinations. By contrast, SIB-200 produces stronger AUC values, particularly for Qwen3, and Belebele lies between these regimes. Figure~\ref{fig:app_error_deciles} summarizes error rates across confidence deciles, while Figure~\ref{fig:app_mmlu_quintiles} examines the MMLU-ProX-Lite failure mode in greater detail.

\begin{table}[t]
\centering
\scriptsize
\setlength{\tabcolsep}{5.6pt}
\begin{tabular}{llccccc}
\toprule
Dataset & Family & En & Bn & Hi & Ur & Mean \\
\midrule
Belebele & Qwen3
& \cellcolor{aucstrong}\textbf{0.785}
& \cellcolor{aucweak}0.618
& \cellcolor{aucmedium}0.653
& \cellcolor{aucweak}0.637
& \cellcolor{aucmedium}0.673 \\
& Gemma 3
& \cellcolor{aucweak}0.604
& \cellcolor{aucweak}0.585
& \cellcolor{aucweak}0.592
& \cellcolor{aucweak}0.577
& \cellcolor{aucweak}0.589 \\
& Llama 3.2
& \cellcolor{aucmedium}0.664
& \cellcolor{aucchance}0.547
& \cellcolor{aucchance}0.539
& \cellcolor{aucchance}0.545
& \cellcolor{aucweak}0.574 \\
\midrule
\multirow{3}{*}{\shortstack{MMLU-\\ProX-Lite}}
& Qwen3
& \cellcolor{aucchance}0.496
& \cellcolor{aucchance}0.494
& \cellcolor{aucchance}0.506
& \cellcolor{aucchance}0.499
& \cellcolor{aucchance}0.499 \\
& Gemma 3
& \cellcolor{aucchance}0.472
& \cellcolor{aucchance}0.495
& \cellcolor{aucchance}0.491
& \cellcolor{aucchance}0.495
& \cellcolor{aucchance}0.488 \\
& Llama 3.2
& \cellcolor{aucweak}0.580
& \cellcolor{aucweak}0.581
& \cellcolor{aucchance}0.534
& \cellcolor{aucchance}0.482
& \cellcolor{aucchance}0.544 \\
\midrule
SIB-200 & Qwen3
& \cellcolor{aucstrong}\textbf{0.840}
& \cellcolor{aucstrong}\textbf{0.802}
& \cellcolor{aucstrong}\textbf{0.830}
& \cellcolor{aucmedium}0.732
& \cellcolor{aucstrong}\textbf{0.801} \\
& Gemma 3
& \cellcolor{aucweak}0.630
& \cellcolor{aucweak}0.636
& \cellcolor{aucweak}0.592
& \cellcolor{aucweak}0.577
& \cellcolor{aucweak}0.609 \\
& Llama 3.2
& \cellcolor{aucstrong}\textbf{0.794}
& \cellcolor{aucmedium}0.660
& \cellcolor{aucmedium}0.677
& \cellcolor{aucmedium}0.683
& \cellcolor{aucmedium}0.704 \\
\bottomrule
\end{tabular}
\caption{Correction-detection AUC by language. Gray denotes weak or near-chance AUC below 0.55, while increasing green intensity indicates stronger correction detection.}
\label{tab:app_auc_language}
\end{table}

\subsection{Correction and Regression Transitions}
\label{app:correction_transitions}

\begin{figure}[htbp]
    \centering
    \includegraphics[width=0.48\textwidth]{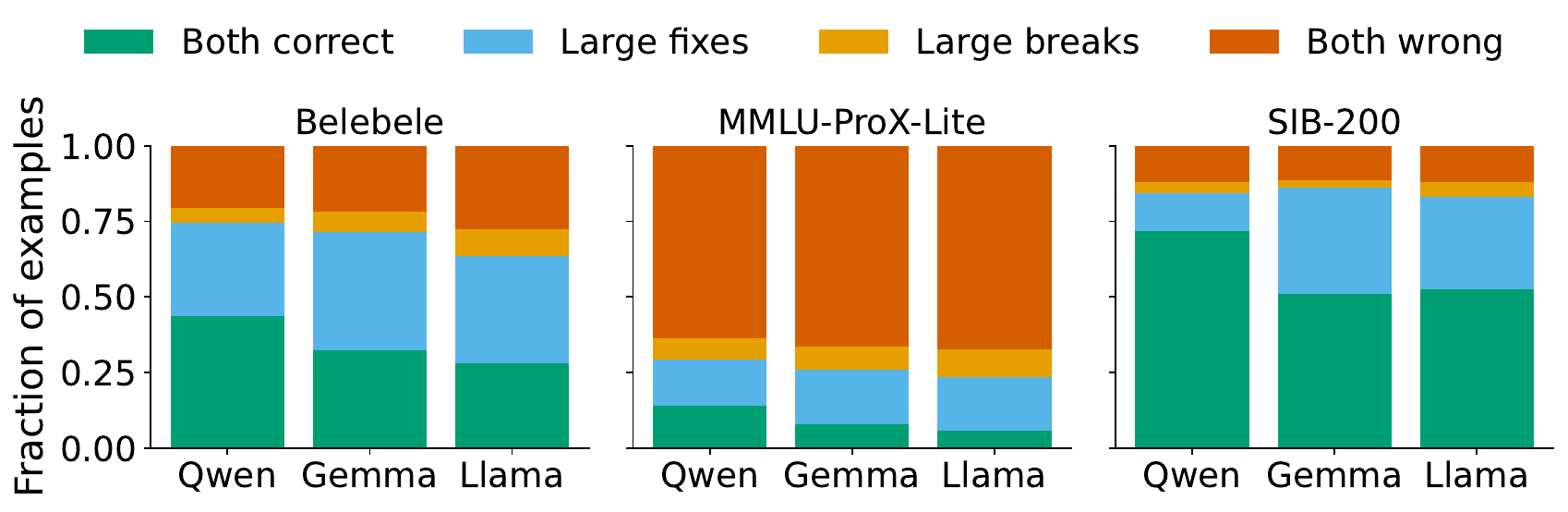}
    \caption{Distribution of small-to-large correctness transitions across datasets and model families.}
    \label{fig:app_transitions}
\end{figure}

Figure~\ref{fig:app_transitions} decomposes each small-to-large comparison into four outcomes. MMLU-ProX-Lite contains the largest proportion of examples missed by both models, ranging from 63.79\% to 67.33\%. Nevertheless, its correction rates remain nontrivial, confirming that the failure of confidence routing is not caused by an absence of correctable examples. Instead, the confidence ordering does not reliably distinguish them.

\subsection{Language-Level Allocation and Outcomes}
\label{app:language_allocation}

Table~\ref{tab:app_language_routing_50} gives the complete language-level allocation and accuracy results for global confidence routing at a nominal 50\% budget. Because examples are ranked globally, individual language escalation rates can deviate considerably from 50\%. The largest difference appears for Qwen3 on SIB-200, where Urdu receives a 60.29\% escalation rate, and English receives 34.80\%.

\begin{table*}[t]
\centering
\scriptsize
\setlength{\tabcolsep}{10.5pt}
\begin{tabular}{llrrrrrr}
\toprule
Family & Policy & Wall time (s) & Throughput & Energy (Wh) & GPU-hours & Time saved & Energy saved \\
\midrule
Qwen & Small only        & $169.1 \pm 3.4$ & 7.10 & 5.23  & 0.094 & 58.9\% & 58.6\% \\
Qwen & Confidence, 25\%  & $290.7 \pm 3.6$ & 4.13 & 8.93  & 0.161 & 29.4\% & 29.4\% \\
Qwen & Equal quota, 25\% & $285.3 \pm 3.5$ & 4.21 & 8.77  & 0.159 & 30.7\% & 30.6\% \\
Qwen & Random, 25\%      & $275.4 \pm 3.5$ & 4.36 & 8.50  & 0.153 & 33.1\% & 32.7\% \\
Qwen & Confidence, 50\%  & $399.0 \pm 2.9$ & 3.01 & 12.26 & 0.222 & 3.1\%  & 3.0\%  \\
Qwen & Equal quota, 50\% & $386.0 \pm 3.5$ & 3.11 & 11.87 & 0.214 & 6.3\%  & 6.1\%  \\
Qwen & Random, 50\%      & $372.3 \pm 3.0$ & 3.22 & 11.46 & 0.207 & 9.6\%  & 9.3\%  \\
Qwen & Large only        & $411.8 \pm 0.1$ & 2.91 & 12.64 & 0.229 & 0.0\%  & 0.0\%  \\
\midrule
Gemma & Small only        & $86.9 \pm 1.1$  & 13.80 & 1.93  & 0.048 & 84.2\% & 88.6\% \\
Gemma & Confidence, 25\%  & $230.4 \pm 0.9$ & 5.21  & 6.34  & 0.128 & 58.2\% & 62.6\% \\
Gemma & Equal quota, 25\% & $230.1 \pm 0.9$ & 5.21  & 6.33  & 0.128 & 58.2\% & 62.6\% \\
Gemma & Random, 25\%      & $222.9 \pm 1.2$ & 5.38  & 6.10  & 0.124 & 59.5\% & 64.0\% \\
Gemma & Confidence, 50\%  & $373.0 \pm 1.2$ & 3.22  & 10.71 & 0.207 & 32.3\% & 36.8\% \\
Gemma & Equal quota, 50\% & $373.7 \pm 1.4$ & 3.21  & 10.70 & 0.208 & 32.1\% & 36.8\% \\
Gemma & Random, 50\%      & $364.8 \pm 1.2$ & 3.29  & 10.46 & 0.203 & 33.7\% & 38.3\% \\
Gemma & Large only        & $550.5 \pm 3.5$ & 2.18  & 16.94 & 0.306 & 0.0\%  & 0.0\%  \\
\midrule
Llama & Small only        & $102.5 \pm 2.6$ & 11.71 & 3.35 & 0.057 & 58.6\% & 57.1\% \\
Llama & Confidence, 25\%  & $179.5 \pm 2.6$ & 6.69  & 5.76 & 0.100 & 27.5\% & 26.3\% \\
Llama & Equal quota, 25\% & $169.1 \pm 2.7$ & 7.10  & 5.44 & 0.094 & 31.6\% & 30.4\% \\
Llama & Random, 25\%      & $168.9 \pm 2.7$ & 7.10  & 5.46 & 0.094 & 31.7\% & 30.2\% \\
Llama & Confidence, 50\%  & $245.1 \pm 2.7$ & 4.90  & 7.86 & 0.136 & 0.9\%  & $-0.5$\% \\
Llama & Equal quota, 50\% & $235.6 \pm 2.4$ & 5.09  & 7.55 & 0.131 & 4.8\%  & 3.5\%  \\
Llama & Random, 50\%      & $229.4 \pm 2.7$ & 5.23  & 7.38 & 0.127 & 7.3\%  & 5.7\%  \\
Llama & Large only        & $247.4 \pm 0.1$ & 4.85  & 7.82 & 0.137 & 0.0\%  & 0.0\%  \\
\bottomrule
\end{tabular}
\caption{Complete system efficiency measurements over 1,200 inputs per family. Wall time is reported as mean $\pm$ standard deviation across three repetitions. Throughput is measured in examples per second. Savings are relative to large-only inference within each family.}
\label{tab:complete_system_efficiency}
\end{table*}

\subsection{Complete System Efficiency Results}
\label{app:system_efficiency_results}

Table~\ref{tab:complete_system_efficiency} reports the complete measurements. At 25\% escalation, confidence routing reduced wall time by 29.4\% for Qwen, 58.2\% for Gemma, and 27.5\% for Llama relative to large-only inference. The corresponding energy reductions were 29.4\%, 62.6\%, and 26.3\%. Throughput increased from 2.91 to 4.13 examples per second for Qwen, from 2.18 to 5.21 for Gemma, and from 4.85 to 6.69 for Llama.

Efficiency gains were substantially smaller at 50\% escalation. Qwen retained wall-time and energy savings of only 3.1\% and 3.0\%, while Llama retained a 0.9\% wall-time reduction and consumed 0.5\% more energy than large-only inference. Gemma retained larger savings of 32.3\% in wall time and 36.8\% in energy because its FP32 large model was considerably more expensive than its FP16 small model.

Confidence routing was sometimes more expensive than equal-quota or random routing at the same escalation rate because the selected subsets differed in computational difficulty. At the 25\% budget, confidence-selected inputs contained an average of 457 tokens for Qwen and 454 for Llama, compared with 415 and 408 tokens under random routing. For Gemma, the corresponding averages were 215 and 200 tokens. Equal escalation rates therefore do not necessarily produce equal latency or energy consumption.

Negative savings indicate that a routed policy consumed more time or energy than the corresponding large-only baseline.

\subsection{Tail Latency}
\label{app:tail_latency}

Table~\ref{tab:tail_latency} distinguishes average efficiency from tail behavior. At 25\% escalation, confidence routing reduced both mean and 95th-percentile latency relative to large-only inference for every family. At 50\%, however, the 95th-percentile latency exceeded large-only inference for Qwen and Llama. Escalated inputs incur both small-model and large-model computation, and confidence tends to select longer inputs for these families. Routing can therefore reduce average computation without guaranteeing lower tail latency.

\begin{table}[htbp]
\centering
\scriptsize
\setlength{\tabcolsep}{9.4pt}
\begin{tabular}{llrrr}
\toprule
Family & Policy & Mean & P50 & P95 \\
\midrule
Qwen & Small only       & 140.4 & 75.3  & 401.2 \\
     & Confidence, 25\% & 241.5 & 111.8 & 882.0 \\
     & Confidence, 50\% & 331.7 & 169.1 & 1151.7 \\
     & Large only       & 342.5 & 199.4 & 1082.2 \\
\midrule
Gemma & Small only       & 72.1  & 70.9  & 80.6 \\
      & Confidence, 25\% & 191.5 & 72.5  & 751.8 \\
      & Confidence, 50\% & 310.2 & 214.8 & 792.0 \\
      & Large only       & 458.3 & 423.1 & 905.4 \\
\midrule
Llama & Small only       & 85.0  & 48.3  & 267.1 \\
      & Confidence, 25\% & 149.0 & 68.7  & 585.7 \\
      & Confidence, 50\% & 203.6 & 102.5 & 795.1 \\
      & Large only       & 205.6 & 127.3 & 630.3 \\
\bottomrule
\end{tabular}
\caption{Mean, median, and 95th-percentile routed latency in milliseconds, averaged across three repetitions.}
\label{tab:tail_latency}
\end{table}

\end{document}